\documentclass[conference]{IEEEtran}
\IEEEoverridecommandlockouts

\usepackage{xspace}
\usepackage{hyperref}
\usepackage{graphicx}
\usepackage{paralist}
\usepackage{amsmath}

\newcommand{\prox}{\ensuremath{\mathop{\mathrm{prox}}}}

\newcommand{\Hc}{\ensuremath{\mathcal{H}}}

\newcommand{\Lc}{\ensuremath{\mathcal{L}}}

\newcommand{\Pc}{\ensuremath{\mathcal{P}}}

\newcommand{\abn}{\ensuremath{\boldsymbol{a}}}

\newcommand{\mbb}{\ensuremath{\boldsymbol{m}}}

\newcommand{\vbn}{\ensuremath{\boldsymbol{v}}}

\newcommand{\wb}{\ensuremath{\boldsymbol{w}}}

\newcommand{\xb}{\ensuremath{\boldsymbol{x}}}

\newcommand{\yb}{\ensuremath{\boldsymbol{y}}}

\newcommand{\zb}{\ensuremath{\boldsymbol{z}}}

\newcommand{\deltab}{\ensuremath{\boldsymbol{\delta}}}

\newcommand{\mub}{\ensuremath{\boldsymbol{\mu}}}

\newcommand{\thetab}{\ensuremath{\boldsymbol{\theta}}}

\newcommand{\Rbb}{\ensuremath{\mathbb{R}}}

\newcommand{\apgdce}{$\text{APGD}_{\text{ce}}$}
\newcommand{\sys}{\textsc{MeanSparse}\xspace}

\usepackage{amssymb}
\newtheorem{definition}{Definition}
\newcommand{\argmin}{\operatornamewithlimits{arg\,min}}
\newcommand{\argmax}{\operatornamewithlimits{arg\,max}}
\graphicspath{{figures/}}
\usepackage{subfigure}
\usepackage{xcolor}
\usepackage{booktabs}

\usepackage{cite}
\usepackage{amsmath,amssymb,amsfonts}
\usepackage{algorithmic}
\usepackage{graphicx}
\usepackage{textcomp}
\usepackage{xcolor}
\def\BibTeX{{\rm B\kern-.05em{\sc i\kern-.025em b}\kern-.08em
    T\kern-.1667em\lower.7ex\hbox{E}\kern-.125emX}}
\begin{document}

\title{MeanSparse: Post-Training Robustness Enhancement Through Mean-Centered Feature Sparsification}

\author{\IEEEauthorblockN{Sajjad Amini, Mohammadreza Teymoorianfard, Shiqing Ma, Amir Houmansadr\\
  \textit{University of Massachusetts Amherst} \\
  \texttt{\{samini, mteymoorianf, shiqingma\}@umass.edu},
  \texttt{amir@cs.umass.edu}}
}

\maketitle

\thispagestyle{plain}
\pagestyle{plain}

\begin{abstract}
We present a simple yet effective method to improve the robustness of both Convolutional and attention-based Neural Networks against adversarial examples by post-processing an adversarially trained model. 
Our technique, \sys,  cascades the activation functions of a trained model with novel operators that sparsify mean-centered feature vectors. 
This is equivalent to reducing feature variations around the mean, and we show that such reduced variations merely affect the model's utility, yet they strongly attenuate the adversarial perturbations and decrease the attacker's success rate.
Our experiments show that, when applied to the top models in the RobustBench leaderboard, \sys achieves a new AutoAttack robust accuracy record of $75.28\%$ (from $73.71\%$), $44.78\%$ (from $42.67\%$) and $62.12\%$ (from $59.56\%$) on CIFAR-10, CIFAR-100 and ImageNet, respectively, in terms of AutoAttack accuracy. 
Code: \url{https://github.com/SPIN-UMass/MeanSparse}
\end{abstract}


\section{Introduction}
\label{sec:intro}

\emph{Adversarial Training} (AT)~\cite{tdlmm18} has become a popular method to defend deep neural networks (DNNs) against adversarial examples~\cite{iposz13, eahgs14}. The core idea involves generating adversarial examples during the training phase and incorporating them into the training dataset. The mere existence of such adversarial examples are due to  features that are non-robust to imperceptible changes~\cite{aeais19}. 
AT attenuates the significance of these non-robust features on model's output, therefore improving robustness.

While AT has demonstrated effectiveness in enhancing model robustness, it faces limitations in generalizability across various methods for generating adversarial examples and suffers from low training efficiency. Various directions have been taken~\cite{wdaxs22, irugr21, fibwr19} to overcome these challenges in order to enhance the robustness provided by AT. In this work, we explore a complementary direction towards improving AT's robustness: The design of activation functions presents a promising, yet underexplored, avenue for improving AT. Smooth Adversarial Training (SAT) has shown that replacing ReLU with a smooth approximation can enhance robustness~\cite{satxt20}, and activation functions with learnable parameters have also demonstrated improved robustness in adversarially trained models~\cite{pafdm22, lafsl23}. This paper aims to further investigate the potential of activation functions in enhancing AT, providing new insights and approaches to bolster model robustness.

In this work, we explore the impact of sparsifying features to enhance robustness against adversarial examples. We introduce the \sys technique, which integrates sparsity into models trained using AT. Our findings indicate that non-robust features persist in the model, albeit with reduced values, even after applying AT. This insight inspired our approach to partially block these remaining non-robust features. To achieve this, we propose imposing sparsity over mean-centered features, denoted by \emph{Mean-based Sparsification}, inspired by sparsity-promoting regularizers commonly used for feature denoising~\cite{idvea06, ksaae06} and feature selection~\cite{rsat96, nfamj10}.

The \sys operator selectively suppresses variations around the mean of feature representations, effectively filtering out non-robust features. For a given feature channel, we compute the mean ($\mu$) and standard deviation ($\sigma$) over the training set. Using a tunable threshold ($ Th=\alpha \sigma$), we block feature values that lie within $\mu \pm Th$, replacing them with the mean value ($\mu$). This operation limits minor perturbations that adversarial attacks often exploit, while preserving the informative structure of features outside this range. For instance, consider a hypothetical feature channel with a mean ($\mu$) of 0.5 and standard deviation ($\sigma$) of 0.2. Setting $\alpha=1$, we block values between 0.3 and 0.7, replacing them with 0.5. This simple mechanism attenuates insignificant variations, as demonstrated in Figure 1 of the paper, where we visualize how the input histogram is transformed. The blocked region corresponds to low-information variations, enhancing robustness by reducing the attacker's exploitable capacity.

\begin{figure*}[t]
    \begin{center}
    \includegraphics[width=0.8\textwidth]{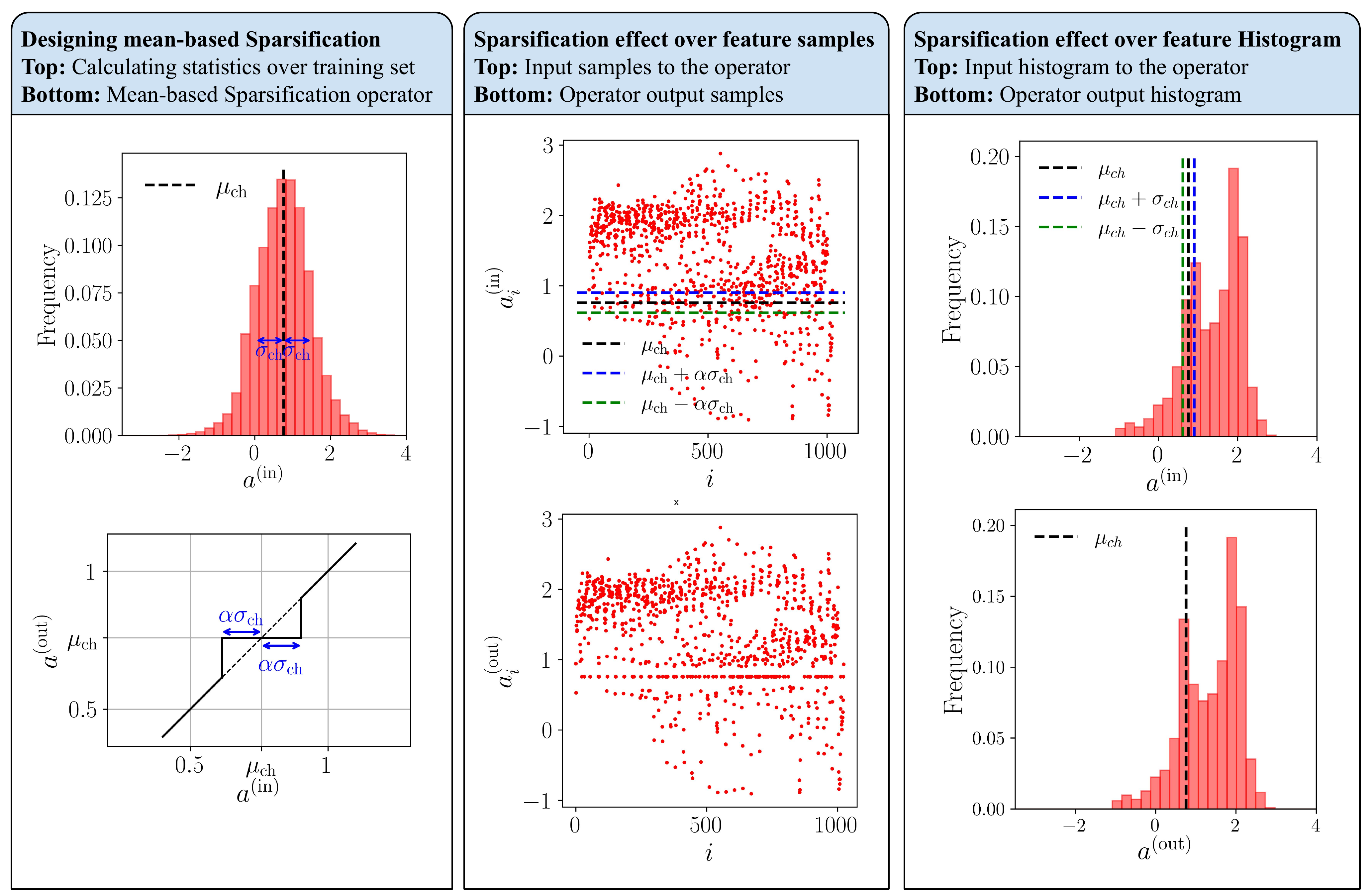}
    \caption{Mean-based sparsification operator used in the \sys technique for hypothetical channel \emph{ch}. The first column represents the design procedue. First, the mean ($\mu_\mathrm{ch}$) and standard deviation ($\sigma_\mathrm{ch}$) are calculated over the training set (top figure). The mean-based sparsification operator is designed with hyper-parameter $\alpha$ which blocks the variations in the $\alpha\sigma_{\mathrm{ch}}$ vicinity of $\mu_\mathrm{ch}$ (bottom figure). The second column represents how mean-based sparsification affects the input features for one test sample (top figure) and generates output features (bottom figure). The effect of mean-based sparsification over the feature histogram is also demonstrated in the third column.}
    \label{fig:pipeline}
    \end{center}
\end{figure*}

Figure \ref{fig:pipeline} illustrates the mean-based sparsification operator, a crucial component of the \sys technique. In the first column, the histogram for a randomly selected channel in the first layer of the RaWideResNet trained on the CIFAR-10 dataset~\cite{rpapx23} is displayed, showing the mean and standard deviation calculated over the CIFAR-10 training set (top figure). The mean-based sparsification operator (bottom figure), parameterized by $\alpha$ blocks variations around the mean. In the second column, the top figure shows the features of a hypothetical channel for a CIFAR-10 test image. The sparsification operator blocks variations between the blue and green dashed lines, producing the output shown in the bottom figure. The third column provides a similar visualization for the input and output histograms of the sparsification operator. Blocking high-probability (low information) variations prevents deterioration of model performance while limiting the attacker's ability to exploit that region.

\sys establishes a new state-of-the-art (SOTA) in robustness in terms of AutoAttack accuracy. When applied to the leading robust models, it improves $\ell_\infty$ AutoAttack accuracy from $73.71\%$ to $75.28\%$ on CIFAR-10, from $42.67\%$ to $44.78\%$ on CIFAR-100, and from $59.56\%$ to $62.12\%$ on ImageNet, while maintaining nearly unchanged clean accuracy. Additionally, when applied to the top model for $\ell_2$ AutoAttack accuracy on CIFAR-10, \sys boosts robust accuracy from $84.97\%$ to $87.28\%$. In all four cases, \sys achieves the highest ranking in terms of AutoAttack accuracy \mbox{\cite{rasca20}}.

In summary, we make the following  contributions:
\begin{compactitem}
    \item {Integrating \sys into the top robust models from the RobustBench benchmark~\mbox{\cite{rasca20}} improved robustness, establishing new records for $\ell_\infty$ AutoAttack accuracy: on CIFAR-10 (from $73.71\%$ to $75.28\%$), CIFAR-100 (from $42.67\%$ to $44.78\%$), and ImageNet (from $59.56\%$ to $62.12\%$). Additionally, new $\ell_2$ AutoAttack accuracy record was set on CIFAR-10 (from $84.97\%$ to $87.28\%$).
    \item The developed \sys technique can be easily integrated into trained models, enhancing their robustness at almost no additional cost.
    \item We identified critical limitations of current activation functions and developed \sys to mitigate these limitations. Our simulations show that regardless of the activation function, \sys improves robustness upon integration.
    \item Through various experiments, we demonstrated that the improved robustness resulting from \sys can be generalized across different model sizes and architecture types.}
\end{compactitem}

\section{Preliminaries}\label{sec:preliminary}

\subsection{Notations}\label{sec:notions}
We use bold lowercase letters to show vectors while the $i$-th element of vector $\xb$ is represented by $x_i$.  Two vector norm metrics are used throughout the paper, including $\ell_0=\Vert\cdot\Vert_0$ (number of non-zero elements) and $\ell_2=\Vert\cdot\Vert_2$ (Euclidean norm). The $\ell_0$ norm can also effectively penalize the variables with small values but its nonsmoothness limits its application. The proximal operator finds the minimizer of function $f$ in the vicinity of input vector $\vbn$ and plays an essential role in our intuition to design the mean-based sparsification operator.
\begin{definition}[Proximal operator \cite{papb14}]\label{def_prox}
    Let $f:\Rbb^n\rightarrow \Rbb\cup \{+\infty\}$ be a proper and lower semi-continuous function. The proximal operator (mapping) $\prox_f:\Rbb^n\rightarrow\Rbb^n$ of $f$ at $\vbn$ is defined as:
    \begin{align}
       \prox_f(\vbn)=\argmin_{\xb}~f(\xb)+\frac{1}{2}\Vert\xb-\vbn\Vert_2^2\label{eq:prox}
    \end{align}
\end{definition}

The definition of proximal operator can be extended to nonsmooth functions such as $\ell_0$ norm. Assume $f(\abn)\triangleq \lambda\Vert \abn\Vert_0$. Then the proximal operator can be calculated as \cite{sare10}:
\begin{align}
    \prox_f(\vbn) = \Hc_{2\lambda}(\vbn),~ \Hc_{\alpha}(v) = 
    \begin{cases}
        v & \vert v \vert > \sqrt{\alpha}\\
        0 & \vert v \vert \leq \sqrt{\alpha}
    \end{cases}
\end{align}
where $\Hc_{\alpha}(\cdot)$ is the element-wise hard-thresholding operator.

\subsection{Related Work}\label{sec:relwk}

Adversarial training, a method to enhance model robustness, encounters high complexity because of the adversarial attack design incorporated during training. The concept of using adversarial samples during training was first introduced in \cite{iposz13} and then AT was introduced in \cite{tdlmm18}. \emph{Fast Gradient Sign Method} (FGSM) is a low-complexity attack design used for AT \cite{eattk18, usslk21, fibwr19, rfsvb20}. This approach while reducing the complexity of AT, cannot generalize to more complex adversarial attacks. On the other hand using \emph{Projected Gradient Descent} with AT, although increasing the attack design complexity due to its iterative nature, can better generalize to other adversarial attacks \cite{tdlmm18, eattk18, otcwm19, ccacl22}. \emph{Curriculum Adversarial Training} gradually increases the complexity of adversarial samples during AT \cite{catcl18}. Other types of adversarial attacks also have been proposed to be used for AT including \emph{Jacobian-based Saliency Map Attack} \cite{oitqh19}, Carlini and Wagner attack \cite{tetcw17, ddcwh18} and attacking Ensemble of methods during AT \cite{iarpx19}, among others.

Different activation functions have been proposed that generally target to improve the generalizability of DNNs \cite{dsrgb11, faacu15, sfarz17, ddihz15, snnlf24, rafll21, swleu18, gelhg18}. Bounded ReLU (BReLU) where the output of ReLU is clipped to avoid adversarial perturbation propagation has shown robustness improvements in standard training scenarios \cite{edazn17}. The effect of symmetric activation functions \cite{stuzg16}, data-dependent activation functions \cite{advwl20}, learnable activation functions \cite{slata21} and activation function quantization \cite{ddnry18} over the robustness of the model in standard training scenario have also been explored.

The performance of different non-parametric activation functions in the AT scenario is explored in \cite{utlgq20} and smooth activation functions demonstrated better robustness. This result is on par with SAT where the authors argued that the ReLU activation function limits the robustness performance of AT due to its non-smooth nature. In other words, the non-smooth nature of ReLU makes the adversarial sample generation harder during training.  Thus they proposed the SAT which replaces ReLU with its smooth approximation and trained using Adversarial training. The results show improved robustness while preserving the accuracy \cite{satxt20}. The effect of non-parametric activation function curvature has also been studied which shows that lower curvature improves the robustness \cite{lcass21}.
 
 Parametric activation functions increase the flexibility of activation functions while the parameters can be learned during the training phase. This flexibility, if designed intelligently, is also shown to be effective in improving robustness with AT. An example of Parametric activation functions is parametric shifted SiLU (PSSILU) \cite{pafdm22}. Similar to Neural Architecture Search (NAS), \emph{Searching for Adversarial Robust Activation Functions} (SARAF) is designed to search over candidate activation functions to maximize the model robustness. Using a meta-heuristic search method, SARAF can handle the intractable complexity of the original search problem \cite{ssfsl23, lafsl23}.

\subsection{Threat Model}\label{abl:threat_model}
In this experiment, we focused on the effectiveness of \sys against two well-studied threat models, $\ell_\infty$ and $\ell_2$. Until now, our focus was on the $\ell_\infty$ threat model, which generally results in lower robust accuracy \cite{rasca20}. On the other hand, $\ell_2$ threat models can focus on a specific part of the image, and this localization of adversarial perturbation may challenge the effectiveness of the \sys technique. In this experiment, we compare the robustness against both $\ell_\infty$ and $\ell_2$ threat models. The model is ResNet-18 with the GELU activation function as in our \emph{Activation Function} experiment in table \ref{tab:activation}. Table \ref{tab:threat} illustrates the clean and APGD accuracy for $\ell_\infty$ and $\ell_2$ threats. As the accuracy for the $\ell_\infty$ threat model increases with increasing $\alpha$, the accuracy for $\ell_2$ also increases, which depicts the effectiveness and generalization of \sys across well-studied threat models. Additionally, the improvement compared to the base model at $\alpha=0.2$ is larger for the $\ell_2$ threat than for the $\ell_\infty$ threat.

\section{Methodology} \label{sec:meth}

From the concept of non-robust features, it is understood that through adversarial training, these features become less informative about the output label \cite{aeais19}. From an information-theoretic point-of-view, we also recognize that as the occurrence probability of a random vector increases, its informational value decreases \cite{eoict06}. Consequently, to identify the less informative features, it is useful to examine the regions of high probability within the feature space. One easily accessible high-probability point is the feature mean. Following this reasoning, we can block minor variations around the mean to eliminate the less informative or equivalently, the non-robust features. We begin this section by designing a regularized optimization objective to block non-robust features. Although this optimization problem is not used in the final \sys technique, its parameter update rule provides valuable insight into the design of the sparsification operator introduced later. Finally, we demonstrate the complete \sys technique.

\subsection{Intuition from Regularized Optimization Objective}

\begin{figure}
    \begin{center}    \centerline{\includegraphics[width=0.4\textwidth]{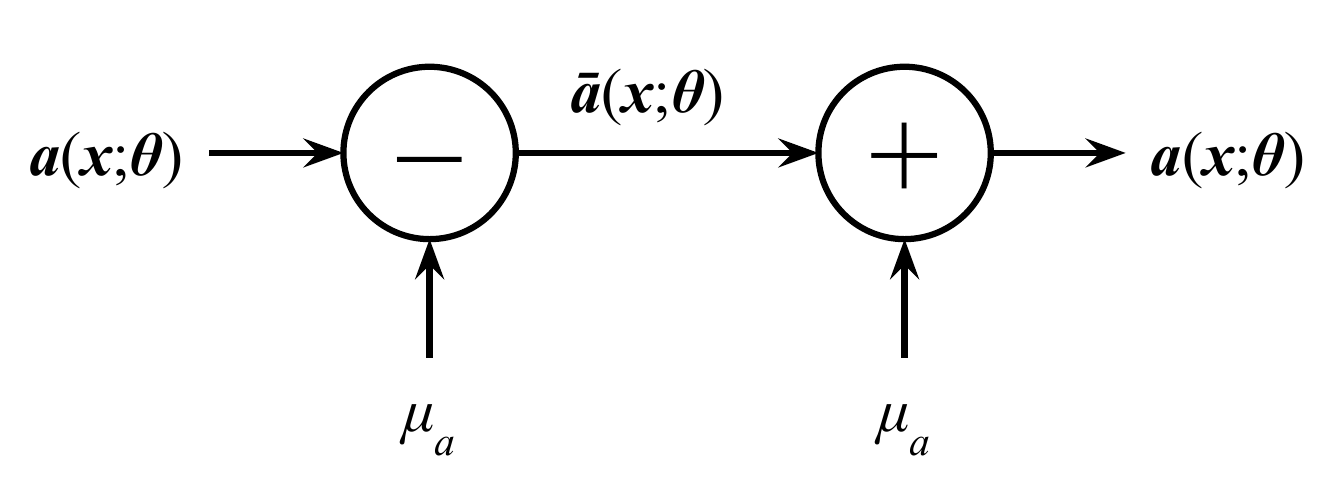}}
    \caption{Mean-centered feature used in regularized optimization problem of \eqref{l0problem}}    
    \label{fig:center}
    \end{center}
\end{figure}

The small changes around the feature mean value at the output of different layers are potentially non-robust features. Thus we start by penalizing those changes using $\ell_0$ norm in an arbitrary layer $l$ for a training sample $\xb$ (extension to the whole dataset and all layers is straightforward). The original training objective can be formulated as:
\begin{align}
    \nonumber\Pc_0:~\thetab_0^\star &= \argmin_{\thetab}~L(\thetab)+\gamma\Vert\overline{\abn}(\xb;\thetab)\Vert_0,\\
    L(\thetab)&\triangleq \Lc(\yb(\xb),\yb^\star;\thetab)\label{l0problem}
\end{align}
where $\thetab$ is the model parameters vector, $\gamma$ is a hyper-parameter, $\overline{\abn}(\xb;\thetab)$ is the centralized feature vector before the activation function in the regularized layer, $\yb(\cdot)$ is the model output and $\yb^\star$ is the target label corresponding to $\xb$.  Throughout the paper, we assume $\mu_a$ to be a fixed vector that is updated using an exponential filter during training, similar to the mean and variance vectors calculated in a batch normalization layer. Figure \ref{fig:center} represents the way mean-centered features for regularization are generated. The optimization problem in \eqref{l0problem} can be solved via penalty method.

To solve problem \eqref{l0problem}, we introduce an approximate version of this problem based on the penalty method (for simplicity we use $\overline{\abn}$ instead of $\overline{\abn}(\xb;\thetab)$):
\begin{align}
\Pc_\lambda:~\thetab^\star_{0,\lambda},\wb^\star = &\argmin_{\thetab,\wb}~L(\thetab)+\gamma\Vert\wb\Vert_0+\frac{1}{2\lambda}\Vert \wb-\overline{\abn}\Vert_2^2\label{sigmamuproblem}
\end{align}
where $\lambda$ is the penalty parameter, $\wb$ is the penalty variable,  and the optimization is with respect to both variables $\thetab$ and $\wb$. Based on the penalty method (PM), we can conclude:
$\lim_{\lambda\rightarrow0}\thetab_{0,\lambda}^\star=\thetab^\star_0$. So to approximate $\thetab_0^\star$, one can solve problem $\Pc_\lambda$ and decrease $\lambda$ during the training. One approach to solve problem $\Pc_\lambda$ is to use block coordinate descent where the $\thetab$ and $\wb$ are updated sequentially \cite{nonw99}. So we have two steps in each training iteration $k$.

\textbf{Step1$\rightarrow$ Calculating $\wb_k$:}
To calculate $\wb_k$, we should solve:
\begin{align}
    \wb_k &=\argmin_{\wb}~\gamma\lambda\Vert\wb\Vert_{0}+\frac{1}{2}\Vert \wb-\overline{\abn}_{k-1}\Vert_2^2\label{proxeq}
\end{align}
The problem in \eqref{proxeq} matches the definition of the proximal operator in \eqref{eq:prox}  and can be solved exactly by the proximal operator for the function $f(\wb)=\gamma\lambda\Vert \wb\Vert_0$. Thus we have:
\begin{align}
    \wb_k = \prox_{f}(\bar{\abn}_{k-1}) = \Hc_{2\lambda\gamma}(\overline{\abn}_{k-1})\label{stepw}
\end{align}

\textbf{Step 2 $\rightarrow$ Calculating $\thetab_k$:}
To calculate $\thetab_k$, we have to solve the following problem:
\begin{align}
    \thetab_{k} = \argmin_{\thetab}~L(\thetab)+\frac{1}{2\lambda}\Vert \wb_k-\overline{\abn}\Vert_2^2\label{eq:theta_update}
\end{align}
while $\wb_k$ is calculated using hard-thresholding operator as in \eqref{stepw}. One step of (stochastic) gradient descent can be used to update $\thetab$ based on \eqref{eq:theta_update}.

\subsection{Mean-centered Feature Sparsification}
Update rule \eqref{eq:theta_update} shows $\overline{\abn}$ approaches $\Hc_{2\lambda\gamma}(\overline{\abn}_{k-1})$ by the second term while the weight for this term is increasing (equivalently $\lambda$ is decreasing) during the training. Based on this intuition, we propose to use the $\Hc_{2\lambda\gamma}(\overline{\abn}_{k-1})$ in the forward propagation of the model. If we add the element-wise mean subtraction and addition in \ref{fig:center} into the hard-thresholding operator, we achieve the curve in the bottom of the first column in \ref{fig:pipeline} assuming $\alpha\sigma_{\mathrm{ch}} = (2\lambda\gamma)^2$ (for simplicity of notation, we use $Th$ instead). This curve represents a sparsification around the feature mean value $\mu_a$. In other words, we have the following element-wise operation over the input $\abn^{(\text{in})}$:
\begin{align*}
    a^{(\text{out})} =
    \begin{cases}
        \mu_a & if~ \vert a^{(\text{out})} - \mu_a \vert \leq~Th \\
        a^{(\text{in})} & if~ \vert a^{(\text{out})} - \mu_a \vert >~Th\\
    \end{cases}
\end{align*}
So, features within the $Th$-vicinity of the feature mean are blocked and the mean value is output, while larger values pass through the Sparsification operator unchanged.

\subsection{MeanSparse Design}
Now we have a mean-centered sparsification operator that can be integrated into the model to block non-robust variations around the mean value. However, there are several challenges and improvements to address, which will be discussed in this section. By applying these enhancements, the final technique is formed.

\subsubsection{Sparsification in Post-processing}
As a general input-output mapping relation, mean-centered feature sparsification can be added to any part of deep learning architecture without changing the dimension. The back-propagation of training signals through this sparsification operator will be zero for inputs in the $Th$-vicinity of the feature mean, which avoids model training since the mean is the place where the highest rate of features lies. Two options can be applied to solve this problem.  

The first approach is to gradually increase the $Th$ value through iterations. At the start of training, $Th$ is set to zero. Thus, sparsification reduces to an identity transformation, which solves the problem of backpropagation. Then, throughout the iterations, $Th$ is increased until it reaches a predefined maximum value. The most challenging problem in this scenario is the selection of a $Th$ scheduler as the model output is not differentiable with respect to $Th$ and $Th$ cannot be updated using gradient. In this scenario, we need to calculate the mean vector for the input feature representation inside the sparsification operator in a similar way to batch normalization layers.

The second solution is to apply sparsification as a post-training step to an adversarially trained model. In this scenario, first, we freeze the model weights and add the mean-centered feature sparsification operator to the model in the predefined positions. Then we need to pass through the complete training set once to calculate the feature mean value. Next, we decide on the value of $Th$. Similar to the previous scenario, this step cannot be done using gradient-based methods as the operator is not differentiable with respect to $Th$. Thus, we can search over a range of $Th$ values and find the best one that results in the highest robustness while maintaining the model's clean accuracy.

The two approaches differ significantly: while training with MeanSparse can improve robustness by influencing learned representations, it requires a carefully designed threshold scheduler to avoid issues like gradient zeroing or instability, particularly in large models, with early experiments showing unstable training and failed convergence. In contrast, the post-training approach integrates more easily, leveraging established model statistics and requiring only a search over alpha values, making it scalable to large models. It has been successfully applied to architectures like Swin-L, achieving a $+2.56\%$ robustness improvement without destabilization. Given the challenges of training MeanSparse in large models and the success of post-training integration, we chose the latter. This involves freezing the model, applying sparsification over mean-centered features before activation functions, calculating the mean value over the training set, and determining the optimal threshold ($Th$) to maximize robustness while maintaining clean accuracy.

\subsubsection{Adaptive Sparsification Using Feature Standard Deviation}
As we need to apply sparsification before all activations in the model, searching over a simple space and finding a suitable set of $Th$ for different activation functions becomes intractable, even in small models. Thus, we design to select the threshold as $Th = \alpha \times \sigma_{\mathrm{a}}$. While we pass through the training set to find the mean value $\mu_{\mathrm{a}}$, we can also determine the variance value $\sigma_a^2$ in parallel and use $Th = \alpha \times \sigma_{\mathrm{a}}$ as the threshold. As a result, for a fixed value of $\alpha$, the blocking vicinity around the mean increases as the variance of the feature increases.

\subsubsection{Per-channel Sparsification}
Throughout the neural network, the input to the activation functions can be considered as 4-dimensional, represented as $N\times C\times H\times W$ ($N$, $C$, $H$ and $W$ represents the batch size, channels, height and width, respectively). To better capture the statistics in the representation, we use per-channel mean $\mu_\mathrm{ch}$ and variance $\sigma_\mathrm{ch}$ in sparsification operator. As a result, the mean and variance vectors are $C$-dimensional vectors. The sparsification operates on each feature channel separately, using the mean and variance of the corresponding channel.

\subsubsection{Complete Pipeline}
The \sys pipeline integrates mean-centered sparsification with three key modifications, placing it before all activation functions with a shared $\alpha$ parameter. Figure \ref{fig:pipeline} shows an example applied to an adversarially trained RaWideResNet on the CIFAR-10 dataset. The first column illustrates a histogram of one channel's features in the first layer, alongside its mean $\mu_{\mathrm{ch}}$ and standard deviation $\sigma_{\mathrm{ch}}$. The corresponding sparsification operator blocks variations near $\alpha\sigma_{\mathrm{ch}}$ around the mean. The second and third columns show the input and output of this operator for a test image from the CIFAR-10 dataset in sample space and histogram. Notably, the feature histogram is bimodal, with one mode aligning with $\mu_{\mathrm{ch}}$, representing uninformative variations blocked by the operator.

By using the \sys presented in this paper, one can enhance the SOTA robust accuracy over CIFAR-10, CIFAR-100 \cite{lmlk09} and ImageNet \cite{ialdd09} datasets, as evidenced by the RobustBench \cite{rasca20} rankings. Through various ablation studies, we also demonstrate the effectiveness of the proposed pipeline in multiple scenarios. This pipeline sheds light on a new approach to improve robustness without compromising clean accuracy.

\section{Experiments} \label{sec:exp}
The proposed pipeline in Figure \ref{fig:pipeline} represents the way one can integrate \sys into different trained models. In this section, we present several experiments to demonstrate the effectiveness of the \sys across different architectures and datasets and in improving the SOTA robustness. The experiments were conducted using an NVIDIA A100 GPU.

\begin{figure*}[tb]
    \centering
    \subfigure[CIFAR-10, $\ell_\infty=\frac{8}{255}$]{
        \includegraphics[width=0.44\textwidth]{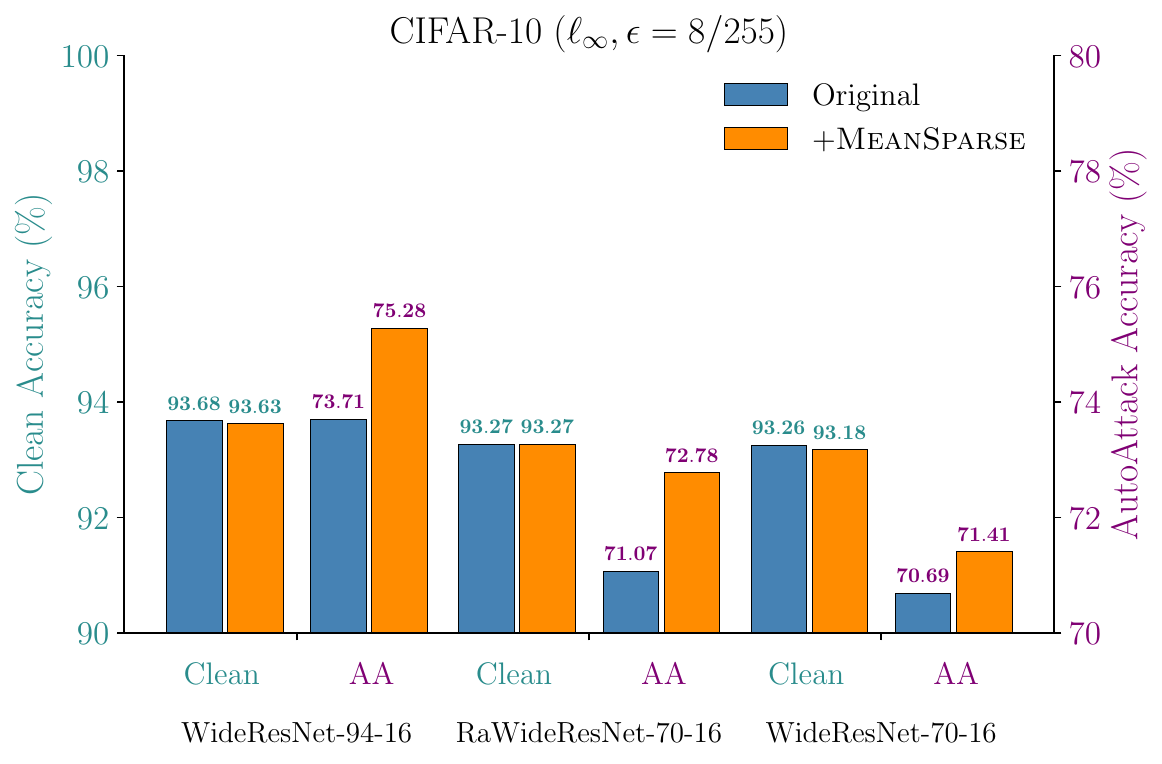}
        \label{fig:bench:sub1}
    }
    \subfigure[CIFAR-10, $\ell_2=0.5$]{
        \includegraphics[width=0.23\textwidth]{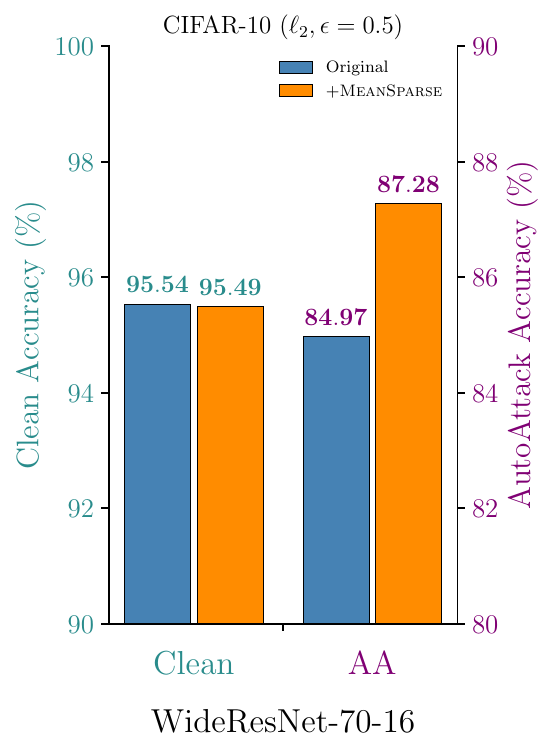}
        \label{fig:bench:sub2}
    }
    \\
    \subfigure[ImageNet , $\ell_\infty=\frac{4}{255}$]{
      \includegraphics[width=0.44\textwidth]{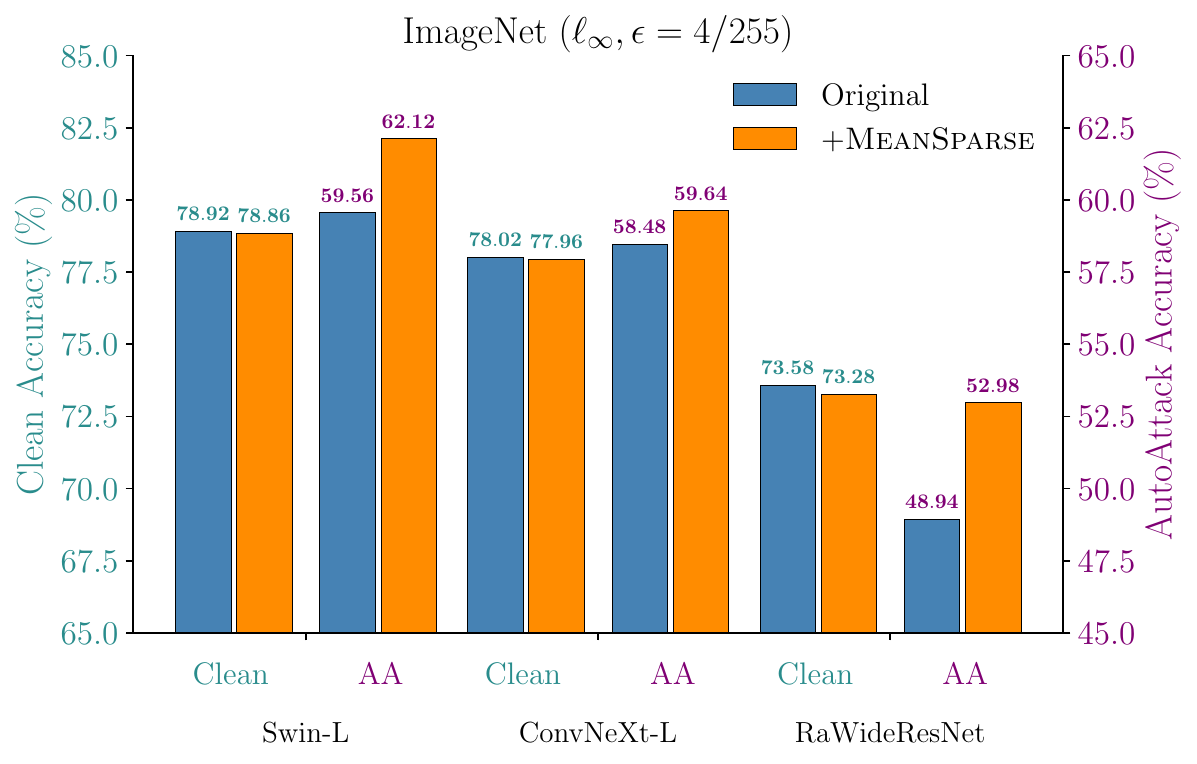}
      \label{fig:bench:sub3}
    }
    \subfigure[CIFAR-100, $\ell_\infty=\frac{8}{255}$]{
        \includegraphics[width=0.23\textwidth]{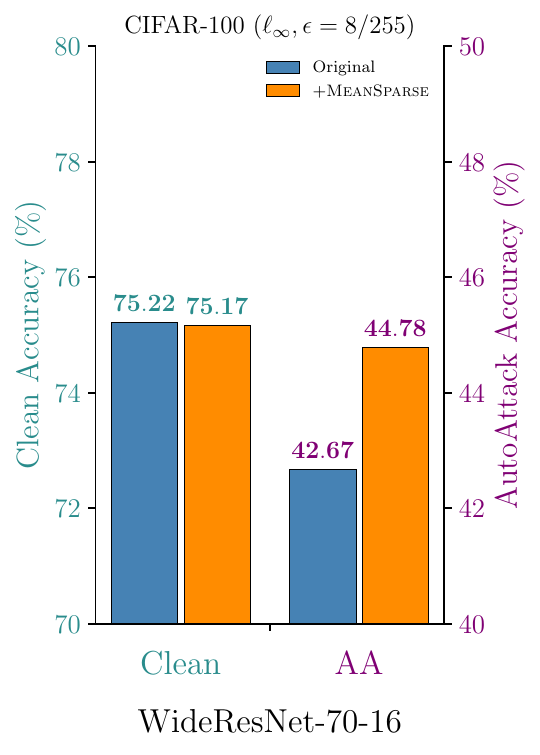}
        \label{fig:bench:sub4}
    }
    \caption{Original models performance along with their performance after integrating with \sys technique. For CIFAR-10 dataset with $\ell_\infty$ attack, we have WideResNet-94-16 (Rank 1) \mbox{\cite{arlbd24}}, RaWideResNet-70-16 (Rank 4) \mbox{\cite{rpapx23}} and WideResNet-70-16 (Rank 5) \mbox{\cite{bdmwp23}} while for $\ell_2$ attack we have   WideResNet-70-16 (Rank 1) \mbox{\cite{bdmwp23}}. For ImageNet dataset with $\ell_\infty$ attack, we have Swin-L (Rank 2) \mbox{\cite{acsld24}}, ConvNeXt-L (Rank 4) \mbox{\cite{acsld23}} and  RaWideResNet (Rank 12) \mbox{\cite{rpapx23}}. For CIFAR-100 dataset with $\ell_\infty$ attack, we have WideResNet-70-16 (Rank 1) \mbox{\cite{bdmwp23}} (all the rankings are based on RobustBench \mbox{\cite{rasca20}})}
    \label{fig:spaf}
\end{figure*}

\subsection{Evaluation Metrics} \label{sec:eval_metric}
Throughout the experiments, we use \emph{Clean} to represent the model accuracy for clean data. \emph{$\text{A-PGD}_{\text{ce}}$}  is used as a robustness measure and represents accuracy after applying Auto PGD with Cross-Entropy (CE) loss. \emph{A-PGD} is used as another robustness measure and represents accuracy after applying Auto-PGD with both CE and Difference of Logits Ratio (DLR) loss metrics. Finally, we use \emph{AA} to represent AutoAttack accuracy. All the robustness metrics can be found in \cite{reoch20}. In our experiments, \sys is applied post-training, eliminating the randomness typically introduced during training. Moreover, results from RobustBench for \emph{$\text{A-PGD}_{\text{ce}}$}, \emph{A-PGD}, and \emph{AA} accuracies are highly consistent, showing negligible variability. Consequently, statistical significance tests are almost zero for all the results reported in this section.

\begin{table*}[!t]
\centering
\footnotesize
\caption{Clean accuracy and accuracy under A-PGD-CE attack \cite{reoch20} (represented by \apgdce) for leading robust models on the CIFAR-10 and ImageNet-1k datasets (\emph{Base} row) and after their integration with the \sys technique (RaWideResNet \cite{rpapx23} and WideResNet \cite{bdmwp23} are leading models in Robustbench ranking for CIFAR-10 dataset and ConvNeXt-L \cite{acsld23} and RaWideResNet \cite{rpapx23} are highly ranked models for ImageNet dataset in Robustbench \cite{rasca20}). The best result in each column is in \textbf{bold} and the threshold selected through the proposed pipeline is in \textcolor{blue}{blue}.}
\renewcommand{\arraystretch}{1.3}
\label{tab:cifar:robustbech:pgd}
\centering
\resizebox{0.85\textwidth}{!}{%
\setlength\tabcolsep{2pt}
\begin{tabular}{cccccccccccccccccc}
    \toprule
    & \multicolumn{11}{c}{CIFAR-10} && \multicolumn{5}{c}{ImageNet-1k} \\
    \cline{2-12}\cline{14-18}
    & \multicolumn{5}{c}{RaWideResNet} &  & \multicolumn{5}{c}{WideResNet}&& \multicolumn{2}{c}{ConvNeXt-L} & & \multicolumn{2}{c}{RaWideResNet}\\
    \cline{2-6}\cline{8-12}\cline{14-15}\cline{17-18}
    & \multicolumn{2}{c}{Train} & & \multicolumn{2}{c}{Test} &  & \multicolumn{2}{c}{Train} & & \multicolumn{2}{c}{Test} && 
    \multicolumn{2}{c}{Test} &&
    \multicolumn{2}{c}{Test}\\
    \cline{2-3}\cline{5-6}\cline{8-9}\cline{11-12}\cline{14-15}\cline{17-18}
    $\alpha$ & Clean & \apgdce && Clean & \apgdce && Clean & \apgdce && Clean & \apgdce && Clean & \apgdce &&Clean & \apgdce\\
    \midrule
    Base & $\mathbf{99.15}$ & $93.44$ && $\mathbf{93.27}$ & $73.87$ && $\mathbf{99.37}$ & $94.69$ && $\mathbf{93.26}$ & $73.45$ &&$78.02$& $60.28$&~   &$\mathbf{73.58}$ & $52.24$\\
    
    $0.15$ & $\mathbf{99.15}$ & $93.69$ && $\mathbf{93.27}$ & $74.40$ && $99.36$ & $\mathbf{94.94}$ && $93.15$ & $73.92$ && $78.00$& $60.62$&    & $73.24$ & $53.38$\\
    
    $0.20$ & $99.13$ & $\mathbf{93.75}$ && $93.26$ & $74.52$ && \textcolor{blue}{$99.33$} & \textcolor{blue}{$94.93$} && \textcolor{blue}{$93.18$} & \textcolor{blue}{$74.11$} && $\mathbf{78.04}$& $60.88$&    & \textcolor{blue}{$73.28$} & \textcolor{blue}{$53.78$}\\
    
    $0.25$ & \textcolor{blue}{$99.10$} & \textcolor{blue}{$93.71$} && \textcolor{blue}{$93.24$} & \textcolor{blue}{$\mathbf{74.74}$} && $99.32$ & $94.85$ && $93.03$ & $\mathbf{74.12}$ && $77.94$& $61.12$&    & $72.82$ & $\mathbf{53.92}$\\
    
    $0.30$ & $99.07$ & $93.49$ && $93.17$ & $74.63$ && $99.29$ & $94.58$ && $92.90$ & $74.04$ && \textcolor{blue}{$77.96$}& \textcolor{blue}{$61.48$}&    & $72.36$ & $53.86$\\
    
    $0.35$ & $98.98$ & $93.12$ && $93.04$ & $74.55$ && $99.23$ & $94.07$ && $92.75$ & $73.72$ && $77.74$& $61.54$&    & $71.74$ & $52.92$\\
    
    $0.40$ & $98.87$ & $92.48$ && $92.88$ & $73.90$ && $99.11$ & $93.27$ && $92.38$ & $73.32$ && $77.56$& $61.84$&    & $69.90$ & $51.68$\\
    
    $0.45$ & $98.70$ & $91.37$ && $92.60$ & $73.36$ && $98.92$ & $91.93$ && $92.08$ & $72.27$ && $77.12$& $62.18$&    & $67.16$ & $48.74$\\
    
    $0.50$ & $98.42$ & $89.76$ && $92.37$ & $72.37$ && $98.66$ & $90.17$ && $91.55$ & $71.10$ && $76.90$& $\mathbf{62.32}$&    & $61.78$ & $43.06$\\
    
    \bottomrule
\end{tabular}
}
\end{table*}

\subsection{Integration to State-of-the-Art Models}
With the introduction of adversarial examples for deep learning models, their reliability was severely questioned \cite{tetcw17}. However, several models have now been developed that present a high level of accuracy while also maintaining acceptable robustness against attacks imperceptible to humans \cite{rasca20}. In this part, we focus on the trained models with leading performance over CIFAR-10, CIFAR-100 \cite{lmlk09} and ImageNet \cite{ialdd09} datasets. We use the trained models and the proposed pipeline to integrate \sys before all the activation functions. Then we evaluate the performance of the resulting models.

Figure \mbox{\ref{fig:spaf}} shows that integrating \sys into the leading  robust models in the RobustBench ranking \mbox{\cite{reoch20}} across various benchmark datasets achieves a new SOTA in AutoAttack accuracy. Figure \mbox{\ref{fig:bench:sub1}} depicts the results for the Clean and AA metrics of the standard and \sys-integrated versions of the WideResNet-94-16 \mbox{\cite{arlbd24}}, RaWideResNet-70-16  \mbox{\cite{rpapx23}} and WideResNet-70-16  \mbox{\cite{bdmwp23}} over CIFAR-10 dataset. The original WideResNet-94-16 \mbox{\cite{arlbd24}} currently holds the SOTA performance in the RobustBench ranking \mbox{\cite{reoch20}} for $\ell_\infty$ untargeted attack. After post-processing with the proposed pipeline, the Clean accuracy drops slightly (from $93.68\%$ to $93.63\%$), while the AA significantly increases (from $73.71\%$ to $75.28\%$), setting a new record for robustness in terms of AA on the CIFAR-10 dataset. Similar results can be obtained for RaWideResNet-70-16  \mbox{\cite{rpapx23}} and WideResNet-70-16  \mbox{\cite{bdmwp23}}. Also, as a result of post-processing the robust model based on the WideResNet-70-16 architecture, the resulting AA accuracy ($71.41\%$) exceeds this value for the robust model based on the RaWideResNet architecture ($71.07\%$), even though WideResNet has a less complex architecture compared to RaWideResNet, which is equipped with the Squeeze-and-Excitation module.

Figure \mbox{\ref{fig:bench:sub2}} represents the result of integrating \sys into the current SOTA robust model (WideResNet-70-16 \mbox{\cite{bdmwp23}}) in terms of AA accuracy with $\ell_2$ distance metric. Similarly, while the clean accuracy is almost unchanged (from $95.54\%$ to $95.49\%$), the AA accuracy is improved (from $84.97\%$ to $87.28\%$).

Figure \mbox{\ref{fig:bench:sub3}} illustrates the clean and AA accuracy for three different models adversarially trained on the ImageNet dataset (Swin-L \mbox{\cite{acsld24}}, ConvNeXt-L  \mbox{\cite{acsld23}} and  RaWideResNet \mbox{\cite{rpapx23}}). By carefully selecting the $\alpha$ value, while the Clean accuracy remains unchanged, the AA accuracy improves significantly. The Swin-L model \mbox{\cite{acsld24}} is currently the second-rank model on the ImageNet dataset. When integrating this model with \sys, while clean accuracy drops slightly (from $78.92\%$ to $78.86\%$), the AA accuracy is increased (from $59.56\%$ to $62.12\%$) setting a new SOTA based on AA accuracy.

Figure \mbox{\ref{fig:bench:sub4}} shows the results of integrating \sys with the current state-of-the-art (SOTA) robust model on the CIFAR-100 dataset (WideResNet-70-16 \mbox{\cite{bdmwp23}}). For this model, the clean and AutoAttack (AA) accuracies are $75.22\%$ and $42.67\%$, respectively. After integrating \sys, these accuracies change to $75.17\%$ and $44.78\%$, establishing a new SOTA robust model with nearly identical clean accuracy.

Table \ref{tab:cifar:robustbech:pgd} presents the results of integrating \sys into RaWideResNet \cite{rpapx23} and WideResNet \cite{bdmwp23} architectures trained on the CIFAR-10 dataset and ConvNeXt-L \cite{acsld23} and RaWideResNet \cite{rpapx23} on the ImageNet dataset using adversarial training. The results are provided for different values of $\alpha$ (note that the threshold equals $\alpha \times \sigma_\mathrm{ch}$). The experiment results over the CIFAR-10 test set depict that for a range of $\alpha$ below $0.25$, the clean accuracy has decreased negligibly or almost unchanged while \apgdce is increasing. Although these results may be attributed to overfitting, the result over the CIFAR-10 training set also represents a similar trend. So an important takeaway from CIFAR-10 results in Table \ref{tab:cifar:robustbech:pgd} is that the variations around the mean value of representation can be considered as a capacity provided by the model for the attackers while its utilization is limited. \sys requires one pass over CIFAR-10 training set to calculate the statistics. The time for this pass is 102 seconds and 95 seconds for the RaWideResNet and WideResNet architectures, respectively. The increase in \apgdce evaluation time after \sys integration is negligible, with only a $2\%$ increase in time.

If this capacity is controlled suitably, then the robustness of the model will be increased (model capacity for attacker will be decreased) while model utilization will be unchanged. So by increasing $\alpha$ from zero, we decrease the attacker capacity while user utilization is unchanged. By increasing $\alpha$ to larger values, we start to decrease user utilization, too. Although there is no rigid margin between these two cases, one can find suitable $\alpha$ values that limit the attacker side while user utilization is unchanged.

Results over ImageNet presented in Table \ref{tab:cifar:robustbech:pgd} show that similar outcomes are achieved when \sys is integrated into models based on the ConvNeXt-L \cite{acsld23} and RaWideResNet \cite{rpapx23} architectures adversarially trained. Note that throughout our experiment over the RaWideResNet \cite{rpapx23} architecture for both CIFAR-10 and ImageNet-1k datasets, we do not apply the sparsification operator inside the Squeeze-and-Excitation module. The time for the \sys pass over the ImageNet-1K training set is 89 minutes for the ConvNeXt-L architecture and 101 minutes for the RaWideResNet architecture. The increase in \apgdce evaluation time after \sys integration is also negligible, with only a $3\%$ increase in time.

\subsection{Activation Function}\label{abl:activation}
\sys is generally appended before the activations function. This questions the effectiveness of \sys for different activation functions. In this experiment, we check the performance of \sys when integrating the models with different activation functions. 
The experiment trains a ResNet-18 model~\cite{ddihz15} on the CIFAR-10 dataset~\cite{lmlk09} for 200 epochs using SGD (initial learning rate: 0.1, weight decay: 0.0005, momentum: 0.9) with a batch size of 256. The learning rate is reduced by a factor of 10 at epochs 100 and 150. Adversarial training is performed using 10-step PGD~\cite{tdlmm18} with respect to $\ell_\infty$ attacks, having a radius of $8/255$ and a step size of 0.0078. The best model is selected based on the highest PGD adversarial accuracy on the test set evaluated at each epoch.

The experiments were conducted using an NVIDIA A100 GPU. Training each ResNet-18 model required approximately 6 hours of computational time on a single A100 GPU. In addition, the evaluation of each trained model took around 20 minutes on the same GPU.

Table \ref{tab:activation} demonstrates the Clean and APGD accuracy metrics for a base ResNet-18 model, denoted by \emph{Base}, with different activation functions and the same model post-processed by the proposed pipeline to integrate MeanSaprse. We use both non-parametric and parametric activation functions, such as PSSiLU \cite{pafdm22}. In our experiment, the GELU activation function, proposed in \cite{gelhg18} and generally used in attention-based architectures, results in the best APGD accuracy. When this model is integrated with \sys, we observe a significant improvement in APGD accuracy ($49.12\%\rightarrow50.37\%$) while the clean accuracy slightly decreases ($84.59\%\rightarrow84.32\%$) for $\alpha=0.2$. Additionally, if we can tolerate $\sim1\%$ reduction in clean accuracy, we can increase APGD accuracy by more than $2\%$. The first important takeaway is the effectiveness of \sys integration with different activations and the inability of the activation function to replicate the role of the \sys technique. The second is that $\alpha=0.2$ consistently leads to large improvements in APDG accuracy with a negligible reduction in clean accuracy for all activation functions. Even for the ReLU activation function, $\alpha=0.2$ increases both Clean and APGD accuracies simultaneously.
We will use the ResNet-18 model trained adversarially with GELU activation over the CIFAR-10 dataset for our future experiments.

\begin{table*}[ht]
\centering
\footnotesize
\caption{Clean and APGD accuracy of ResNet-18 model with different activation functions over CIFAR-10 test set before (Base) and after integration with \sys technique for different value of $\alpha$}
\renewcommand{\arraystretch}{1.3}
\label{tab:activation}
\centering
\resizebox{0.75\textwidth}{!}{
\setlength\tabcolsep{2pt}
\begin{tabular}{cccccccccccccccccc} 
    \toprule
    \multicolumn{2}{c}{ReLU}&&\multicolumn{2}{c}{ELU} && \multicolumn{2}{c}{GELU} && \multicolumn{2}{c}{SiLU} && \multicolumn{2}{c}{PSiLU} && \multicolumn{2}{c}{PSSiLU}\\
    \cline{2-3} \cline{5-6} \cline{8-9}\cline{11-12}\cline{14-15}\cline{17-18}
    $\alpha$ & Clean & APGD &&Clean & APGD &&Clean & APGD && Clean & APGD && Clean & APGD && Clean & APGD \\ \midrule
    Base& $83.59$ & $47.77$ && $\mathbf{81.65}$ & $46.65$ && $84.59$ &$49.12$ &&$83.05$ & $48.54$ && $\mathbf{84.80}$ & $48.02$ && $83.95$ & $48.03$\\
    
    0.05&$83.58$ & $47.89$ && $81.64$ & $47.47$ && $\mathbf{84.63}$ & $49.25$ &&$83.10$ & $48.70$ && $\mathbf{84.80}$ & $48.11$ && $\mathbf{83.97}$ & $48.22$\\
    
    0.1& $83.59$ & $48.25$ &&$81.58$ & $48.69$ &&$84.58$ & $49.57$ &&$83.09$ & $49.05$ && $84.72$ & $48.58$ && $83.95$ & $48.61$\\
    
    0.15& $83.57$ & $48.60$ && $81.53$ & $49.22$ &&$84.42$ & $50.00$ &&$\mathbf{83.11}$ & $49.73$ && $84.66$ & $48.88$ && $83.93$ & $49.05$\\
    
    0.2& $\mathbf{83.71}$ & $48.88$  && $81.53$ & $\mathbf{49.49}$ &&$84.32$ & $50.37$ &&$83.01$ & $50.01$ && $84.69$ & $49.07$ && $83.79$ & $49.29$\\
    
    0.25& $83.46$ & $\mathbf{49.30}$ && $81.56$ & $49.14$ &&$84.18$ & $50.80$ &&$83.00$ &$50.59$ && $84.61$ & $\mathbf{49.36}$ && $83.75$ & $\mathbf{49.32}$\\
    
    0.3& $83.48$ & $49.24$ && $81.20$ & $48.66$ &&$83.95$ & $51.13$ &&$82.78$ &$\mathbf{50.81}$ && $84.27$ & $49.35$ && $83.81$ & $49.19$\\
    
    0.35& $83.13$ & $49.27$ && $80.34$ & $47.60$ &&$83.51$ & $\mathbf{51.36}$ &&$82.49$ &$50.79$ && $83.76$ & $49.25$ && $83.55$ & $49.11$\\
    
    0.4& $82.49$ & $48.87$ && $78.77$ & $46.10$ &&$82.85$ & $51.25$ &&$81.34$ &$49.97$ && $83.20$ & $48.76$ && $82.69$ & $48.63$\\
    
    \bottomrule
    \end{tabular}
}
\end{table*}

\begin{table}[h]
\centering
\footnotesize
\caption{Clean and APGD accuracy of ResNet-18 model with GELU activation function over CIFAR-10 test set before (Base) and after integration with \sys technique for different value of $\alpha$ and different  threat models from RobustBesnch \cite{reoch20}}
\renewcommand{\arraystretch}{1.3}
\label{tab:threat}
\centering
\setlength\tabcolsep{2pt}
\resizebox{0.3\textwidth}{!}{
\begin{tabular}{cccccccc} 
    \toprule
    &&&&\multicolumn{1}{c}{$\ell_\infty, \epsilon = \frac{8}{255}$} &&\multicolumn{1}{c}{$\ell_2 , \epsilon = \frac{128}{255}$ } \\
    \cline{3-3} \cline{5-5} \cline{7-7}
    $\alpha$ && Clean && APGD && APGD \\
    \midrule
    
    Base &&$84.59$ && $49.12$ && $59.98$ \\
    $0.05$ &&$\mathbf{84.63}$ && $49.25$ && $60.23$ \\
    $0.10$ &&$84.58$ && $49.57$ && $60.53$ \\
    $0.15$ &&$84.42$ && $50.00$ && $60.87$ \\
    $0.20$ &&$84.32$ && $50.37$ && $61.36$ \\
    $0.25$ &&$84.18$ && $50.80$ && $61.60$ \\
    $0.30$ &&$83.95$ && $51.13$ && $61.73$ \\
    $0.35$ &&$83.51$ && $\mathbf{51.36}$ && $\mathbf{61.74}$\\
    $0.40$ &&$82.85$ && $51.25$ && $\mathbf{61.74}$ \\
    
    \bottomrule
    \end{tabular}
}
\end{table}

\subsection{Blackbox Attacks}\label{abl:black}
The \sys operation masks the gradient during backward propagation when the input $a^{\mathrm{in}}$ lies within $\alpha\sigma_{\mathrm{ch}}$ of the channel mean $\mu_{\mathrm{ch}}$ (see the bottom plot in the first column of Figure \mbox{\ref{fig:pipeline}}). An interesting direction to explore is the performance of black-box attacks that do not rely on gradients, both before and after integrating \sys. For this experiment, we select the Square Attack \mbox{\cite{saaac20}}. Table \mbox{\ref{tab:black}} shows the clean and robust accuracy for a ResNet-18 model with GELU activation before (Base) and after integration with \sys (robust accuracy is denoted by \mbox{\textit{Square}}, representing the accuracy under Square Attack). The results indicate a similar improvement in robust accuracy on par with APGD, while clean accuracy remains largely unaffected across a wide range of $\alpha$ values, from $0.05$ to $0.2$.

\begin{table}[t]
\caption{Clean and Square accuracy (Square accuracy in the \mbox{\textit{Square}} column represents the accuracy after applying square attack \mbox{\cite{saaac20}}) of ResNet-18 model with GELU activation function over CIFAR-10 test set before (Base) and after integration with \sys technique for different value of $\alpha$}
\renewcommand{\arraystretch}{1.3}
\label{tab:black}
\centering
\setlength\tabcolsep{2pt}
\resizebox{0.15\textwidth}{!}{
\begin{tabular}{cccccccc} 
    \toprule
    $\alpha$ && Clean & Square \\
    \midrule
    Base &&$84.59$ & $57.37$ \\
    $0.05$ &&$\mathbf{84.63}$ & $57.66$ \\
    $0.10$ &&$84.58$ & $58.19$  \\
    $0.15$ &&$84.42$ & $59.31$  \\
    $0.20$ &&$84.32$ & $60.30$  \\
    $0.25$ &&$84.18$ & $60.67$  \\
    $0.30$ &&$83.95$ & $61.32$  \\
    $0.35$ &&$83.51$ & $\mathbf{61.52}$ \\
    $0.40$ &&$82.85$ & $61.49$ \\
    \bottomrule
    \end{tabular}
}
\end{table}

\subsection{Adaptive Attack}\label{sec:adaptive}
In this experiment, we investigate the potential for the attacker to enhance the attack by utilizing the knowledge that \sys is employed to robustify the model. We propose a formulation to adapt the original PGD attack, incorporating the integration of \sys to create a more powerful attack. In this scenario, we assume the attacker has complete access to the locations, mean, and variance vectors of the integrated \sys operator throughout the model. Let $\xb$ represent the input image and $\{\zb_i\}$ the corresponding representations at the input of the \sys modules in the model. When a perturbation $\deltab$ is added to the input ($\xb + \deltab$), it results in new representations $\{\zb_i^\delta\}$. In the adaptive PGD attack, we solve the following optimization problem to determine the perturbation $\deltab$:
\begin{align}
    \nonumber&\argmax_{\deltab} \quad \mathrm{CE}(f(\xb + \deltab), y) - \lambda \sum_i \Vert \mbb_i \odot (\zb_i-\zb_i^\delta)\Vert_p^q\\
    &\text{subject to} \quad \|\deltab\|_\infty \leq \epsilon, \quad \xb + \deltab \in [0, 1]^D\label{form:adaptive}
\end{align}
where $\mathrm{CE}(\cdot, \cdot)$ is the cross-entropy loss, $f$ is the model, $y$ is the true label, $\lambda$ is a hyper-parameter, $i$ indexes different \sys modules in the model, $\odot$ is the element-wise multiplication operator, $\epsilon$ controls the perturbation imperceptibility via the $\ell_\infty$ norm, and $D$ is the dimension of the input image. $\mbb_i$ represents a mask with the same dimensions as $\zb_i$, identifying the locations in $\zb_i$ affected by the \sys module. For a given $\zb$, the corresponding $\mbb$ can be computed element-wise as:
\begin{align*}
    m_j = 
    \begin{cases}
        1 & \text{if~~}\mu_{\mathrm{ch}} - \alpha\sigma_{\mathrm{ch}} \leq z_j \leq \mu_{\mathrm{ch}} + \alpha\sigma_{\mathrm{ch}}\\
        0 & \text{Otherwise}
    \end{cases}, j=1, 2, \ldots, K
\end{align*}
where $\mu_{\mathrm{ch}}$ and $\sigma_{\mathrm{ch}}$ are the statistics of the \sys module, and $K$ represents the representation dimension. Thus, whenever a feature in the input of \sys is affected by this module for the input image $\xb$, the corresponding mask value becomes $1$; otherwise, it is $0$. As a result, the second term in the objective function of the adaptive PGD attack \mbox{\eqref{form:adaptive}} penalizes perturbations that lead to changes in the features affected by \sys. In the ideal case, the perturbation is designed such that none of the \sys modules can block its effect in their input (we know that the changes in the input of the $i$-th \sys module are $\zb_i - \zb_i^d$). Thus, the attacker can bypass the \sys, which we denote as the Adaptive-$(p, q)$ attack.

To test the adaptive attack, we solve the optimization problem \mbox{\eqref{form:adaptive}} using gradient ascent with a step size of $\mu = \frac{1}{255}$, a number of iterations $N = 50$, and $\epsilon = \frac{8}{255}$. When $\lambda = 0$, problem \mbox{\eqref{form:adaptive}} reduces to the well-known PGD-50 attack, while for larger values of $\lambda$, we obtain an adaptive attack. We conducted experiments using this adaptive attack on a ResNet-18 model with a GELU activation function, which was adversarially trained on the CIFAR-10 dataset. Table \mbox{\ref{tab:adaptive:attack}} compares the results of the PGD-50 attack with the Adaptive-$(1,1)$ and Adaptive-$(2,2)$ attacks before (Base) and after integration ($\alpha = 0.20$ and $\alpha = 0.35$) with \sys. For the base model, all the attacks are equivalent, as $\mbb_i$ is a zero matrix for all \sys modules. As $\alpha$ increases, the accuracy after applying our adaptive attack decreases compared to the PGD-50 attack, demonstrating the effectiveness of our adaptation. While the adapted attack is more powerful than the original PGD-50 attack, the improvement gap in robustness resulting from \sys is still significant. For the base model, the robust accuracy is $51.56\%$, while the worst accuracies are $53.56\%$ and $55.26\%$ for $\alpha = 0.20$ and $\alpha = 0.35$, respectively, indicating a clear improvement in robustness, while the clean accuracy remains almost unchanged.

\begin{table}[h]
\footnotesize
\caption{Clean, PGD-50 and Adaptive Attack accuracy for a ResNet-18 model with GELU activation over CIFAR-10 dataset before (Base) and after integration with \sys}
\renewcommand{\arraystretch}{1.3}
\label{tab:adaptive:attack}
\centering
\setlength\tabcolsep{2pt}
\resizebox{0.4\textwidth}{!}{
\begin{tabular}{cccccccccc} 
    \toprule
    $\alpha$ && Clean & PGD-50 & Adaptive-$(1,1)$ & Adaptive-$(2,2)$ \\
    \midrule
    Base &&$\mathbf{84.59}$ & $51.56$ & $51.56$ & $51.56$ \\
    $0.20$ &&$84.32$ & $53.89$ & $53.56$ & $53.58$ \\
    $0.35$ &&$83.51$ & $\mathbf{55.50}$ & $\mathbf{55.36}$ & $\mathbf{55.26}$ \\
    \bottomrule
    \end{tabular}
}
\end{table}

\subsection{Adversarial Training}\label{abl:adv_tr}
\sys is designed to integrate with models trained using adversarial training. In previous experiments, PGD \cite{tdlmm18} was used for adversarial training. In this section, we compare the PGD results with those of TRADES \cite{zhang2019theoretically} for adversarial training. For TRADES adversarial training, we set $\beta = 0.6$. For the other properties of adversarial training, we adopt the parameters specified in Section \ref{tab:activation}. Table \ref{tab:type} compares the results for base adversarially trained models and models integrated with \sys. For TRADES, \sys can still improve the APGD accuracy while maintaining clean accuracy. This table demonstrates overlapping information gathered around the mean feature values for both PGD and TRADES as adversarial training methods. In both methods, the information around the feature mean with a value of $\alpha\simeq 0.2$ is almost uninformative for clean data but can be utilized by attackers. Thus, the performance of \sys represents generalizability across both PGD and TRADES adversarial training methods.

\begin{table}[tb]
\footnotesize
\caption{Clean and APGD accuracy of ResNet-18 model with GELU activation function over CIFAR-10 test set before (Base) and after integration with \sys technique for different value of $\alpha$ and adversarial training based on PGD \cite{tdlmm18} and TRADES \cite{zhang2019theoretically}}
\renewcommand{\arraystretch}{1.3}
\label{tab:type}
\centering
\setlength\tabcolsep{2pt}
\resizebox{0.3\textwidth}{!}{
\begin{tabular}{cccccccc} 
    \toprule
    &&\multicolumn{2}{c}{PGD} &&\multicolumn{2}{c}{TRADES}\\
    \cline{3-4} \cline{6-7}
    $\alpha$ && Clean & APGD && Clean & APGD \\
    \midrule
    Base &&$84.59$ & $49.12$ && $83.01$ & $47.03$ \\
    $0.05$ &&$\mathbf{84.63}$ & $49.25$ && $83.04$ & $47.16$ \\
    $0.10$ &&$84.58$ & $49.57$ && $83.09$ & $47.61$ \\
    $0.15$ &&$84.42$ & $50.00$ && $\mathbf{83.11}$ & $48.08$ \\
    $0.20$ &&$84.32$ & $50.37$ && $83.08$ & $48.64$ \\
    $0.25$ &&$84.18$ & $50.80$ && $83.02$ & $48.92$ \\
    $0.30$ &&$83.95$ & $51.13$ && $82.92$ & $\mathbf{49.34}$ \\
    $0.35$ &&$83.51$ & $\mathbf{51.36}$ && $82.62$ & $49.24$\\
    $0.40$ &&$82.85$ & $51.25$ && $82.20$ & $48.76$\\
    \bottomrule
    \end{tabular}
}
\end{table}

\subsection{Standard Training}\label{abl:std}
In the previous experiments, we explore the effect of two different adversarial training methods on performance improvement by \sys. In this experiment, we test \sys integration over a standard trained model. Table \mbox{\ref{tab:std:attack:power}} represents the APGD accuracy over a ResNet-18 model with standard training for $\epsilon$ values starting from $1/255$ to $8/255$. The first column ($\alpha=0$) is the base model performance while the next two columns ($\alpha = 0.2$ and $\alpha=0.35$) represent the models integrated with \sys. In all cases, the attacker can fool the model with imperceptible perturbation ($\epsilon\leq 4/255$) which necessitates the integration of \sys with adversarially trained models.

\begin{table}[tb]
\footnotesize
\caption{APGD accuracy of standard trained ResNet-18 model with GELU activation function over CIFAR-10 test set before (Base denoted by $\alpha=0$ in the table) and after the integration of \sys technique for different attack power $\epsilon$ and two different $\alpha$ values ($0.20$ and $0.35$). Clean accuracy is provided next to each $\alpha$ value.}
\renewcommand{\arraystretch}{1.3}
\label{tab:std:attack:power}
\centering
\setlength\tabcolsep{2pt}
\resizebox{0.35\textwidth}{!}{
\begin{tabular}{ccccccc} 
    \toprule
    &&\multicolumn{1}{c}{0  ($94.36$)}&& \multicolumn{1}{c}{0.2 ($94.20$)} && \multicolumn{1}{c}{0.35 ($93.61$)}\\
    \cline{3-3} \cline{5-5} \cline{7-7}
    $\epsilon$ && APGD && APGD && APGD\\ \midrule
    $1/255$ && $49.09$ && $52.06$ && $56.36$\\
    $2/255$ && $4.86$ && $7.69$ && $14.52$\\
    $3/255$ && $0.07$ & & $0.24$ && $1.33$\\
    $4/255$ && $0.0$ & & $0.01$ && $0.03$\\
    $5/255$ && $0.0$ & & $0.0$ && $0.01$\\
    $6/255$ && $0.0$ & & $0.0$ && $0.01$\\
    $7/255$ && $0.0$ & & $0.0$ && $0.01$\\
    $8/255$ && $0.0$ & & $0.0$ && $0.01$\\
    \bottomrule
    \end{tabular}
}
\end{table}

\subsection{Feature Centerization Type}\label{abl:center}
The \sys technique applies sparsification over mean-centered features across each channel in the input tensor to the activation function. In this section, we inspect the importance of the reference for feature centralization before sparsification in our proposed technique. For this purpose, we compare three cases where the reference for centering features is the channel mean ($\mub_{ch}$), zero, and the mean over all features in all channels ($\mu$). For a fair comparison across all cases, we select the sparsification threshold equal to $\alpha \times \sigma_{ch}$, where $\sigma_{ch}$ emphasizes the channel-wise calculation standard deviation. Table \ref{tab:center:type} represents the result for different options of reference. When using zero as the reference for feature centralization, the improvement in APGD is higher than using the mean as the reference, but on the other hand, the utilization of the model also decreases. One challenging problem is the high rate of clean accuracy reduction when using zero as the reference, which may cause a huge reduction if not selected properly. When using the mean of all channels as the reference, the results in terms of both clean and APGD accuracy metrics underperform compared to the case where the channel mean is used as the reference. The main takeaway from the table is the importance of using channel-wise statistics to maintain utility while improving robustness.

\begin{table}[h]
\footnotesize
\caption{Clean and APGD accuracy of ResNet-18 model with GELU activation function over CIFAR-10 test set before (Base) and after the application of \sys technique for different references for feature centralization}
\renewcommand{\arraystretch}{1.3}
\label{tab:center:type}
\centering
\resizebox{0.45\textwidth}{!}{%
\setlength\tabcolsep{2pt}
\begin{tabular}{ccccccccccc} 
    \toprule
    &&\multicolumn{2}{c}{zero} &&\multicolumn{2}{c}{$\mu$} &&\multicolumn{2}{c}{$\boldsymbol{\mu}_{ch}$} \\
    \cline{3-4} \cline{6-7} \cline{9-10}
    $\alpha$ && Clean & APGD && Clean & APGD && Clean & APGD \\
    \midrule
    $Base$ && $\mathbf{84.59}$ & $49.12$ && $\mathbf{84.59}$ & $49.12$ &&$84.59$ & $49.12$  \\
    $0.05$ && $84.58$ &$49.32$ && $84.58$ & $49.22$ &&$\mathbf{84.63}$ & $49.25$ \\
    $0.1$ && $84.50$ & $49.63$ && $84.52$ & $49.45$ &&$84.58$ & $49.57$ \\
    $0.15$ && $84.30$ &$50.26$ && $84.36$ & $49.79$ &&$84.42$ & $50.00$ \\
    $0.2$ && $84.06$ &$50.98$ && $84.13$ & $50.23$ &&$84.32$ & $50.37$ \\
    $0.25$ && $83.55$ &$51.58$ && $83.83$ & $50.73$ &&$84.18$ & $50.80$ \\
    $0.3$ && $82.81$ &$\mathbf{52.08}$ && $83.27$ & $\mathbf{50.87}$ &&$83.95$ & $51.13$ \\
    $0.35$ && $81.68$ &$51.95$ && $82.52$ & $50.61$ &&$83.51$ & $\mathbf{51.36}$ \\
    $0.4$ && $79.84$ &$51.72$ && $81.09$ & $49.93$ &&$82.85$ & $51.25$ \\
    
    \bottomrule
    \end{tabular}
}
\end{table}

\subsection{Attack Power}\label{abl:power}
In the previous experiments, the attack power was set to a fixed value of $8/255$ for $\ell_\infty$ attack threat which is generally the threshold of imperceptibility \cite{reoch20}. In this experiment, we explore the effectiveness of the \sys technique for a wide range of attack powers. Table \ref{tab:attack:power} represents the APGD accuracy for $\epsilon$ values starting from $1/255$ to $16/255$. For $\alpha=0.2$, the accuracy is smaller than the base model ($\alpha=0$) for $1/255$ and $2/255$ values of $\epsilon$, which is generally due to the reduction of clean accuracy after applying \sys (clean accuracy reduces to $84.32$ from $84.59$). For $\epsilon$ values larger than $2/255$, applying the \sys technique persistently improves the APGD accuracy. The same trend is also observed for $\alpha=0.35$ except for the fact that after $\epsilon=4/255$, we have persistent improvements over the base model.

\begin{table}[h]
\footnotesize
\caption{APGD accuracy of ResNet-18 model with GELU activation function over CIFAR-10 test set before (Base denoted by $\alpha=0$ in the table) and after the application of \sys technique for different attack power $\epsilon$ and two different $\alpha$ values ($0.20$ and $0.35$). Clean accuracy is provided next to each $\alpha$ value.}
\renewcommand{\arraystretch}{1.3}
\label{tab:attack:power}
\centering
\resizebox{0.38\textwidth}{!}{
\setlength\tabcolsep{2pt}
\begin{tabular}{ccccccc} 
    \toprule
    &&\multicolumn{1}{c}{0  ($84.59$)}&& \multicolumn{1}{c}{0.2 ($84.32$)} && \multicolumn{1}{c}{0.35 ($83.51$)}\\
    \cline{3-3} \cline{5-5} \cline{7-7}
    $\epsilon$ && APGD && APGD && APGD\\ \midrule
    $1/255$ && $81.10$ && $80.97$ && $79.60$\\
    $2/255$ && $77.72$ && $77.60$ && $76.47$\\
    $3/255$ && $73.37$ & & $73.54$ && $72.64$\\
    $4/255$ && $68.88$ & & $69.08$ && $68.80$\\
    $5/255$ && $64.31$ & & $64.70$ && $64.48$\\
    $6/255$ && $58.95$ & & $59.92$ && $60.15$\\
    $7/255$ && $54.11$ & &$55.15$ && $55.71$\\
    $8/255$ && $49.12$ & &$50.37$ && $51.36$\\
    $9/255$ && $42.77$ & &$44.41$ && $45.72$\\
    $10/255$ && $37.48$ & &$39.17$ && $40.77$\\
    $11/255$ && $32.62$ && $34.20$ && $36.04$\\
    $12/255$ && $28.12$ && $29.77$ && $31.50$\\
    $13/255$ && $23.77$ & &$25.44$ &&$27.18$\\
    $14/255$ && $19.42$ & &$20.76$ && $22.77$\\
    $15/255$ && $15.51$ & &$17.01$ && $19.09$\\
    $16/255$ && $11.97$ & &$13.41$ && $15.44$\\
    \bottomrule
    \end{tabular}
}
\end{table}

\subsection{Applying to Specific Activation Functions}\label{abl:pattern}
The \sys technique is designed to sparsify mean-centered inputs to all activation functions in a model. In this experiment, we will focus on applying sparsification over a selected subset of activation functions. For this purpose, we index the activation functions in the ResNet-18 model. This model has 8 residual blocks, each with 2 activation functions: one in the \emph{Main Path} and one after the result of the main path is added to the residual connection, totaling 16 activation functions. Additionally, this architecture has one activation function before the residual block. So, in total, we have 17 indices. Index 0 is the initial activation function. The odd indices are related to the activation function in the main path, and the even indices correspond to activation functions after the addition of the residual connection to the main path output. We select two scenarios. In the first scenario, denoted by \emph{Single Activation}, we present the results when \sys is applied to only the corresponding row-indexed activation function, while \emph{Cumulative} depicts the result when \sys is applied to the activation functions with indices less than or equal to the index represented in each row. 

Table \ref{tab:imagenet:robustbech:pgd} represents the results for both \emph{Single Activation} and \emph{Cumulative} scenarios. One shared remark is that the starting layers are more effective in improving robustness. We can attribute this to the abstraction level in different layers of the ResNet-18 model. In the case of Single Activation, when we apply the \sys technique to larger indices, we observe that not only does the robust accuracy decrease, but the clean accuracy increases. This indicates that for even better performance, we can use different $\alpha$ values for activation functions based on their depth level. Note that for the current version of the \sys technique, we use a shared $\alpha$ for all activation functions.

Another aspect worth inspecting in this experiment is how the application of the \sys technique to activation functions in the main path and after the addition of residual data to the main path result affects robustness in terms of APGD accuracy. So we consider two cases in Table \ref{tab:connection}. In the first case, \sys is applied to all the activation functions in the main path of the ResNet block, denoted by \emph{Main Path}. In the second case, \sys is applied to all the activation functions after the addition of residual data to the output of the main path, denoted by \emph{After Addition}. This table reveals an important result. While the application of \sys to both types of activation functions contributes to the robustness of the model, the application of \sys to the main flow of the model (\emph{After Addition}) is more responsible for the improvements in the ranges where clean accuracy is almost unchanged ($\alpha \simeq 0.2$).

\begin{table}[h]
\caption{Clean and APGD accuracy of ResNet-18 model with GELU activation function over CIFAR-10 test set after the application of \sys technique for \emph{Single Activation} and \emph{Cumulative } cases. Single Activation represents the results when \sys is applied to only the index activation function, and Cumulative Activation represents the results when \sys is applied to the index activation function and all previous activation functions (for the based model the Clean and APGD accuracy are $84.59\%$ and $49.12\%$, respectively).}
\renewcommand{\arraystretch}{1.3}
\label{tab:imagenet:robustbech:pgd}
\centering
\resizebox{0.5\textwidth}{!}{
\setlength\tabcolsep{2pt}
\begin{tabular}{ccccccccccccc} 
    \toprule
    & \multicolumn{5}{c}{Single Activation} && \multicolumn{5}{c}{Cumulative} \\
    \cline{2-6}\cline{8-12}
    & \multicolumn{2}{c}{0.2}&& \multicolumn{2}{c}{0.35} & &\multicolumn{2}{c}{0.2}& &\multicolumn{2}{c}{0.35} \\ \cline{2-3} \cline{5-6}\cline{8-9}\cline{11-12}
    Index & Clean &APGD && Clean &APGD && Clean &APGD && Clean &APGD\\ \midrule
    $0$ & $84.52$ & $\mathbf{49.34}$ & &$84.21$ & $49.19$ &&$84.52$ & $49.37$ && $84.21$ & $49.19$\\
    $1$ & $84.56$ & $49.19$ && $84.44$ & $49.09$ &&$\mathbf{84.53}$ & $49.38$ && $\mathbf{84.31}$ & $49.35$\\
    $2$ & $84.46$ & $49.21$ && $84.45$ & $49.20$ &&$84.49$ & $49.37$ && $83.98$ & $49.29$\\
    $3$ & $84.50$ & $49.15$ && $84.33$ & $49.19$ &&$84.45$ & $49.52$ && $83.72$ & $49.69$\\
    $4$ & $84.48$ & $49.21$ && $84.26$ & $\mathbf{49.26}$ &&$84.41$ & $49.67$ && $83.47$ & $50.02$\\
    $5$ & $84.54$ & $49.24$ && $84.36$ & $49.25$ &&$84.38$ & $49.86$ && $83.14$ & $50.29$\\
    $6$ & $84.60$ & $49.11$ && $84.58$ & $49.21$ &&$84.34$ & $49.90$ && $83.21$ & $50.50$\\
    $7$ & $84.51$ & $49.13$ && $84.45$ & $49.16$ &&$84.32$ & $49.99$ && $83.02$ & $50.65$\\
    $8$ & $84.58$ & $49.19$ && $84.59$ & $49.21$ &&$84.34$ & $50.12$ && $83.05$ & $50.86$\\
    $9$ & $84.60$ & $49.12$ && $84.53$ & $49.02$ &&$84.35$ & $50.32$ && $82.96$ & $51.09$\\
    $10$ & $84.59$ & $49.06$ && $84.59$ & $49.07$ &&$84.34$ & $50.35$ && $83.12$ & $51.15$\\
    $11$ & $84.60$ & $49.06$ && $84.56$ & $49.05$ &&$84.38$ & $50.33$ && $83.02$ & $51.23$\\
    $12$ & $84.62$ & $49.09$ && $84.72$ & $49.05$ &&$84.37$ & $50.27$ && $83.09$ & $51.23$\\
    $13$ & $84.59$ & $49.13$ && $84.63$ & $49.09$ &&$84.38$ & $50.37$ && $83.14$ & $51.20$\\
    $14$ & $84.60$ & $49.17$ && $84.63$ & $49.12$ &&$84.40$ & $50.34$ &&$83.26$ & $\mathbf{51.41}$\\
    $15$ & $84.57$ & $49.15$ && $84.69$ & $49.10$ &&$84.36$ & $\mathbf{50.38}$ && $83.30$ & $51.37$\\
    $16$ & $\mathbf{84.63}$ & $49.09$ && $\mathbf{84.75}$ & $49.01$ &&$84.32$ & $50.37$ && $83.51$ & $51.36$\\
    \bottomrule
    \end{tabular}
}
\end{table}

\begin{table*}[h]
\footnotesize
\caption{Clean and PGD-50 \mbox{\cite{tdlmm18}} accuracy of various RobustBench-ranked models \mbox{\cite{rasca20}}, both before (Base) and after integration with the \sys technique (models tested here includes WideResNet-94-16 \mbox{\cite{arlbd24}} and RaWideResNet-70-16 \mbox{\cite{rpapx23}} over CIFAR-10 dataset, WideResNet-70-16 \mbox{\cite{bdmwp23}} over CIFAR-100 dataset and Swin-L \mbox{\cite{acsld24}} and ConvNeXt-L \mbox{\cite{acsld23}} over ImageNet}
\renewcommand{\arraystretch}{1.3}
\label{tab:attack_type}
\centering
\resizebox{0.8\textwidth}{!}{%
\setlength\tabcolsep{2pt}
\begin{tabular}{cccccccccccccccccccccc}
    \toprule
    & \multicolumn{5}{c}{CIFAR-10} && \multicolumn{3}{c}{CIFAR-100} & \multicolumn{5}{c}{ImageNet} \\
    \cline{2-6}\cline{8-9}\cline{11-15} 
    & \multicolumn{2}{c}{WideResNet-94-16} && \multicolumn{2}{c}{RaWideResNet-70-16} && \multicolumn{2}{c}{WideResNet-70-16} && \multicolumn{2}{c}{Swin-L} && \multicolumn{2}{c}{ConvNeXt-L}\\
    \cline{2-3}\cline{5-6}\cline{8-9}\cline{11-12}\cline{14-15} 
     & Clean & PGD-50 && Clean & PGD-50 && Clean & PGD-50 && Clean & PGD-50 && Clean & PGD-50 \\
    \midrule
    Base & $\mathbf{93.68}$ & $76.36$ && $\mathbf{93.27}$ & $74.01$ && $\mathbf{75.23}$ & $48.35$ && $\mathbf{78.92}$ & $61.42$ && $\mathbf{78.02}$ & $60.62$\\
    
    \sys & $93.63$ & $\mathbf{78.21}$ && $93.24$ & $\mathbf{75.17}$ && $75.17$ & $\mathbf{50.64}$ && $78.88$ & $\mathbf{61.78}$ && $77.96$ & $\mathbf{61.78}$\\
    
    \bottomrule
\end{tabular}}
\end{table*}

\begin{table*}[h]
  \centering
  \caption{Relative MAE between subset and full-dataset feature statistics for various robust models.}
  \label{tab:subset}
  \renewcommand{\arraystretch}{1.1}
  \setlength{\tabcolsep}{4pt}
  \resizebox{0.9\textwidth}{!}{%
  \begin{tabular}{c|cc|cc|cc|cc}
    \toprule
    & \multicolumn{2}{c|}{WRN-70-16 (CIFAR-10, $\ell_2$) \cite{bdmwp23}} 
    & \multicolumn{2}{c|}{WRN-94-16 (CIFAR-10, $\ell_\infty$) \cite{arlbd24}} 
    & \multicolumn{2}{c|}{WRN-70-16 (CIFAR-100) \cite{bdmwp23}} 
    & \multicolumn{2}{c}{Swin-L (ImageNet) \cite{acsld24}} 
    \\ \cmidrule{2-9}
    \textbf{Subset (\%)} 
          & \textbf{Mean} & \textbf{Std} 
          & \textbf{Mean} & \textbf{Std}
          & \textbf{Mean} & \textbf{Std}
          & \textbf{Mean} & \textbf{Std} \\
    \midrule
     10   & 0.1476 & 0.0033 & 0.1286 & 0.0038 & 0.3481 & 0.0044 & 0.0228 & 0.0275 \\
     25   & 0.0471 & 0.0015 & 0.0707 & 0.0017 & 0.0930 & 0.0019 & 0.0119 & 0.0274 \\
     50   & 0.0382 & 0.0010 & 0.0520 & 0.0010 & 0.1000 & 0.0010 & 0.0037 & 0.0272 \\
     75   & 0.0228 & 0.0005 & 0.0205 & 0.0006 & 0.0784 & 0.0006 & 0.0046 & 0.0272 \\
    \bottomrule
  \end{tabular}%
  }
\end{table*}

\begin{table}[h]
\footnotesize
\caption{Clean and APGD accuracy of ResNet-18 model with GELU activation function over CIFAR-10 test set before (Base denoted by $\alpha=0$ in the table) and after the application of \sys technique for \emph{Main Path} (application of \sys to activation functions in the main path of residual block) and \emph{After Addition} (application of \sys to activation functions operating after addition of main path result and residual data) cases.}
\renewcommand{\arraystretch}{1.3}
\label{tab:connection}
\centering
\setlength\tabcolsep{2pt}
\resizebox{0.3\textwidth}{!}{
\begin{tabular}{cccccccc} 
    \toprule
    &&\multicolumn{2}{c}{Main Path} &&\multicolumn{2}{c}{After Addition}\\
    \cline{3-4} \cline{6-7}
    $\alpha$ && Clean & APGD && Clean & APGD \\
    \midrule
    Base &&$\mathbf{84.59}$ & $49.12$ && $84.59$ & $49.12$ \\
    $0.05$ &&$84.59$ & $49.20$ && $\mathbf{84.62}$ & $49.25$ \\
    $0.10$ &&$84.59$ & $49.31$ && $84.59$ & $49.43$ \\
    $0.15$ &&$84.50$ & $49.43$ && $84.49$ & $49.53$ \\
    $0.20$ &&$84.40$ & $49.63$ && $84.47$ & $49.66$ \\
    $0.25$ &&$84.26$ & $49.68$ && $84.46$ & $\mathbf{49.93}$ \\
    $0.30$ &&$83.94$ & $49.88$ && $84.27$ & $49.90$ \\
    $0.35$ &&$83.66$ & $\mathbf{49.94}$ && $84.15$ & $49.73$\\
    $0.40$ &&$82.96$ & $49.91$ && $83.67$ & $49.24$ \\
    \bottomrule
    \end{tabular}
}
\end{table}

\subsection{Attack Type}\label{abl:pgd50}
Figure \mbox{\ref{fig:spaf}} demonstrates the improvement in robustness of leading models in RobustBench in terms of AutoAttack accuracy. In this experiment, we use PGD-50 \mbox{\cite{tdlmm18}} for robustness measurement. Table \mbox{\ref{tab:attack_type}} represents the Clean and PGD-50 accuracy values for different models before (denoted by Base) and after integration with \sys (denoted by \sys) over CIFAR-10, CIFAR-100 and ImageNet datasets. The results again confirm the improvements in Figure \mbox{\ref{fig:spaf}} as t integration with \sys increases the PGD-50 accuracy in all models.

\subsection{Robustness with Subset-Based Statistics}

A practical advantage of \sys lies in its minimal data requirements at post-training. Instead of accessing the full training dataset, the method only requires feature statistics (mean and variance), which can be reliably estimated from subsets of the data. Table~\ref{tab:subset} presents the approximation quality when statistics are computed from 10\%, 25\%, 50\%, and 75\% of the training set, evaluated by Relative Mean Absolute Error (MAE). We observe that Relative MAE values below 10\% already yield accurate estimates, while values below 5\% indicate close alignment with full-data statistics. For instance, with only 25\% of CIFAR-10 data, the WideResNet-70-16 model achieves a Relative MAE of approximately 5\%, leading to natural and robust accuracies of \emph{95.49\% and 87.31\%}, respectively. These results closely match the full-data performance of 95.49\% and 87.28\%, demonstrating that MeanSparse maintains reliability even when computed with partial datasets. This finding highlights the scalability of the method to scenarios where full training data access is restricted due to privacy, storage, or efficiency constraints.

\section{Limitations} \label{sec:limitation}
The \sys technique proposed in this paper is a post-training method that is useful on top of adversarially trained models. Using \sys for models trained in a standard way does not lead to robustness improvement as we expect based on our intuition introduced in this paper. Additionally, while integrating a model with \sys shows clear improvements against both white-box and black-box non-adaptive attacks, white-box attacks that ignore the MeanSparse transformation during backpropagation and use the identity function can impact \sys's effectiveness.

\section{Conclusions}
Adversarial training, a widely accepted defense against evasion attacks, attenuates non-robust feature extractors in a model. Although adversarial training (AT) is effective, this paper demonstrates that there is still an easily accessible capacity for attackers to exploit. We introduce the \sys technique, which partially blocks this capacity in the model available to attackers and enhances adversarial robustness. This technique, easily integrable after training, improves robustness without compromising clean accuracy. By integrating this technique into trained models, we achieve a new record in adversarial robustness in terms of AutoAttack accuracy on CIFAR-10 for both $\ell_\infty$ and $\ell_2$ untargeted attacks, as well as on CIFAR-100 and ImageNet for $\ell_\infty$ untargeted attacks. 

\bibliographystyle{IEEEtran}
\bibliography{satml}


\end{document}